\documentclass[letterpaper, 10 pt, conference]{ieeeconf}

\IEEEoverridecommandlockouts                              

\usepackage{times}
\usepackage{multicol}
\usepackage{graphicx}
\usepackage{subcaption}

\usepackage{graphics}
\usepackage{epsfig}
\usepackage{mathptmx}
\usepackage{times}
\usepackage{amsmath} 
\usepackage{amssymb} 
\usepackage{mathtools}
\usepackage{xcolor}
\usepackage{url}
\usepackage{multicol}
\usepackage{booktabs}
\usepackage{soul}
\usepackage{xcolor}
\usepackage[hang,flushmargin]{footmisc}
\usepackage{cite}
\usepackage{balance}

\renewcommand{\vec}[1]{\mathbf{#1}}
\newcommand\norm[1]{\left\lVert#1\right\rVert}

\title{\LARGE \bf
Joint Inference of Kinematic and Force Trajectories\\with Visuo-Tactile Sensing
}

\author{Alexander (Sasha) Lambert$^{1,3}$, Mustafa Mukadam$^{1,3}$, Balakumar Sundaralingam$^{2,3}$, \\ Nathan Ratliff$^{3}$, Byron Boots$^{1,3}$, and Dieter Fox$^{3,4}$
\thanks{$^{1}$Georgia Institute of Technology, Robot Learning Lab, USA
}%
\thanks{$^{2}$University of Utah, Robotics Center and the School of Computing, USA
}%
\thanks{$^{3}$NVIDIA, USA
}%
\thanks{$^{4}$University of Washington, Paul G.~Allen School of Computer Science \& Engineering, Seattle, WA, USA
}%
\thanks{Email: {\tt\small alambert6@gatech.edu}}
}

\begin{document}

\setstcolor{red}

\maketitle
\thispagestyle{empty}
\pagestyle{empty}

\begin{abstract}

To perform complex tasks, robots must be able to interact with and manipulate their surroundings. One of the key challenges in accomplishing this is robust state estimation during physical interactions, where the state involves not only the robot and the object being manipulated, but also the state of the contact itself.
In this work, within the context of planar pushing, we extend previous inference-based approaches to state estimation in several ways. We estimate the robot, object, and the contact state on multiple manipulation platforms configured with a vision-based articulated model tracker, and either a biomimetic tactile sensor or a force-torque sensor.
We show how to fuse raw measurements from the tracker and tactile sensors to jointly estimate the trajectory of the kinematic states and the forces in the system via probabilistic inference on factor graphs, in both batch and incremental settings.
We perform several benchmarks with our framework and show how performance is affected by incorporating various geometric and physics based constraints, occluding vision sensors, or injecting noise in tactile sensors. We also compare with prior work on multiple datasets and demonstrate that our approach can effectively optimize over multi-modal sensor data and reduce uncertainty to find better state estimates.

\end{abstract}


\section{Introduction  \& Related Work}

Manipulation is a difficult problem, complicated by the challenge of robustly estimating the state of the robot's interaction with the environment. Parameters such as the contact point and the force vector applied at that point, can be very hard to robustly estimate. These parameters are generally partially observable and must be inferred from noisy information obtained via coarse visual or depth sensors and highly sensitive but difficult to interpret tactile sensors.

For instance, in the case of ``in-hand" manipulation problems, where a held object is often partially occluded by an end-effector, tactile sensing offers an additional modality that can be exploited to estimate the pose of the object \cite{schmidt2015depth}.

Vision and tactile sensors have been used to localize an object within a grasp using a gradient-based optimization approach~\cite{bimbo2013combining}. This has been extended to incorporate constraints like signed-distance field penalties and kinematic priors~\cite{schmidt2015depth}. However, the former is deterministic and the latter handles uncertainty only per time-step, which is insufficient since sensors can be highly noisy and sensitive. Particle filtering-based approaches have been proposed that can infer the latent belief state from bi-modal and noisy sensory data, to estimate the object pose for two-dimensional grasps~\cite{zhang2012application} and online localization of a grasped object~\cite{chalon2013online}. These approaches are often limited in scope. For example, \cite{chalon2013online} uses vision to only initialize the object pose and later relies purely on contact information and dynamics models. In general, particle filtering based methods also suffer from practical limitations like computational complexity, mode collapse, and particle depletions in tightly constrained state spaces.

\begin{figure}[!t]
	\centering
	\vspace{2mm}
	\begin{subfigure}[b]{0.45\textwidth}
		\includegraphics[width=\textwidth]{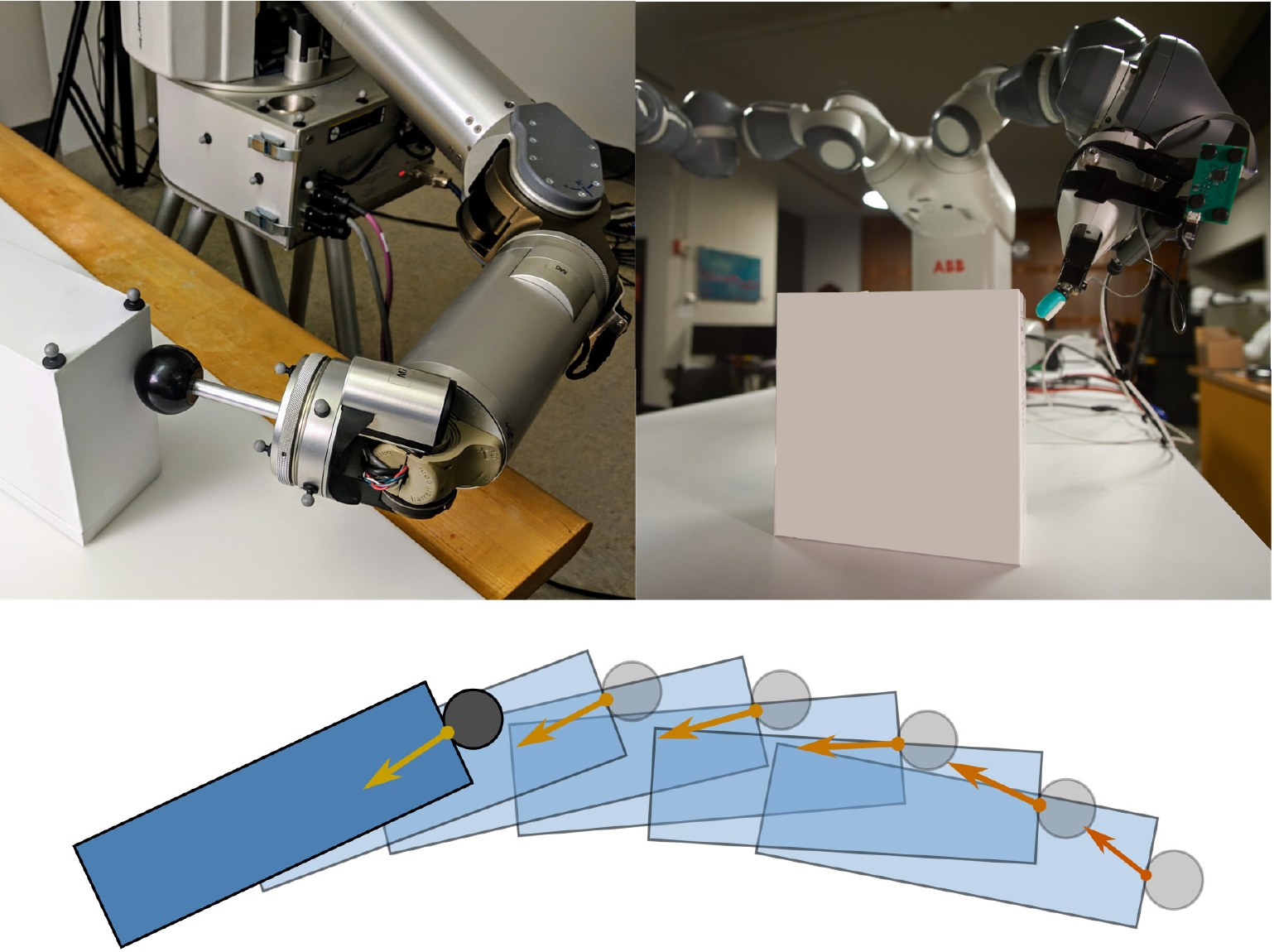}
		\label{realyumi}
	\end{subfigure} 
	\caption{\small Tracking contact dynamics: (Top-left) Pushing probe with Force-Torque sensor on the WAM arm. (Top-right) Yumi robot with mounted biomimetic tactile sensor. (Bottom) Optimized kinematic and force trajectories on a pushed object.}
	\label{intro_img}
	\vspace{-6mm}
\end{figure}

Beyond manipulation, sate estimation is a classic problem in robotics. For example, Simultaneous Localization and Mapping (SLAM) has been studied for many decades, and many efficient tools have been developed to address noisy multi-modal sensor fusion in these domains~\cite{montemerlo2002fastslam,thrun2006graph,dellaert2006square}.
One of the more successful tools, the smoothing and mapping (SAM) framework~\cite{dellaert2006square}, uses factor graphs to perform inference and exploits the underlying sparsity of the estimation problem to efficiently find locally optimal distributions of latent state variables over temporal sequences. This technique offers the desired combination of being computationally fast while accounting for uncertainty over time, and has been recently incorporated into motion planning~\cite{mukadam2018continuous,Mukadam-RSS-17}.

This framework has also been explored for estimation during manipulation~\cite{yu2015shape,yu2017realtime,yu2018realtime}. In particular, Yu et al.~\cite{yu2017realtime} formulate a factor graph of planar pushing interaction (for non-prehensile and underactuated object manipulation) using a simplified dynamics model, with both visual object-pose and force-torque measurements and show improved pose recovery over trajectory histories compared to single-step filtering techniques.
However, the scope of~\cite{yu2017realtime} is limited to the use of a purpose-built system, equipped with small-diameter pushing-rods kept at a vertical orientation, allowing for high-fidelity contact-point estimation. A fiducial-based tracking system is also used. Such high precision measurements are impractical in a realistic setting.

In this work, we extend the capabilities of such factor graph inference frameworks in several ways to perform planar pushing tasks in real world settings. 
We extend the representation to incorporate various geometric and physics-based constraints alongside multi-modal information from vision and tactile sensors. We perform ablation benchmarks to show the benefits of including such constraints, and benchmarks where the vision is occluded or the tactile sensors are very noisy, using data from on our own generalized systems. We conduct our tests on two systems, a dual-arm ABB Yumi manipulator equipped a gel-based Syntouch Biotac tactile sensor~\cite{biotac} and a Barrett WAM arm equipped with a pushing probe end effector mounted with a force torque sensor (see Fig.\ref{intro_img}). Both of these systems are also set up with a vision-based articulated tracking system that leverages a depth camera, joint encoders, and contact-point estimates~\cite{schmidt2015depth}.

Through inference, we jointly estimate the history of not only object poses, and end-effector poses, but also, contact points, and applied force vectors. Estimating contact points and applied force vectors can be very useful in tractable dynamics models to predict future states and can be beneficial to contact-rich planning and control for manipulation~\cite{hogan2016feedback}.

With our experiments, we show that we can contend with a range of multi-modal noisy sensor data and perform efficient inference in batch and incremental settings to provide high-fidelity and consistent state estimates.

\section{Dynamics of Planar Pushing}

In this section, we review the dynamics model for pushing on planar surfaces.
The quasi-static approximation of this model is used in the next section to describe the motion model of the pushed object within the factor graph for estimation.

Given an object of mass $m$ being pushed with an applied force $f$, we can describe the planar dynamics of the rigid body through the primary equations of motion
\begin{equation}
f + f_\mu = m\ddot{x}_{CM}, \quad \tau + \tau_\mu = I_{CM}\omega
\label{eqn_motion_moment}
\end{equation}
where $x_{CM}$ is the object position measured at the center-of-mass (CM), $\omega$ the angular velocity of the object frame, $I_{CM}$ the moment of inertia, and $f_{\mu}$ the linear frictional force. The applied and frictional moments are defined as $\tau = {x}_{CM} \times f$ and $\tau_{\mu} = {x}_{CM} \times f_{\mu}$ respectively.

We can estimate the frictional loads on the object by considering the contribution of each point on the support area $A$ of the object \cite{yu2015shape}. The friction force $f_\mu$ and corresponding moment $\tau_{\mu}$ is found by integrating Coulomb's law across the contact region of the object with the surface
\begin{equation}
\small
f_{\mu} =-\mu_s \int_{A} \frac{v(r)}{|v(r)|}P(r)dA, \quad \tau_{\mu} =-\mu_s \int_{A} r \times \frac{v(r)}{|v(r)|}P(r)dA \label{eqn_frict}
\end{equation}
where $v(r)$ denotes the linear velocity at a point $r$ in area $A$, and $P(r)$ the pressure distribution. The coefficient of friction is assumed to be uniform across the support area.

For pusher trajectories that are executed at near-constant speeds, inertial forces can be considered negligible. The push is then said to be quasi-static, where the applied force is large enough to overcome friction and maintain a velocity, but is insufficient to impart an acceleration~\cite{lynch1992manipulation}. Then, the applied force $f$ must lie on the limit surface. This surface is defined in $(f_x,f_y,\tau)$ space and encloses all loads under which the object would remain stationary~\cite{mason1986mechanics}. It can be approximated as an ellipsoid with principal semi-axes $f_{max}$ and $\tau_{max}$~\cite{lee1991fixture}
\begin{equation}
\left(\frac{f_x}{f_{max}}\right)^2 + \left(\frac{f_y}{f_{max}}\right)^2 + \left(\frac{\tau}{\tau_{max}}\right)^2 = 1
\end{equation}
where $f_{max}=\mu_s f_n$, and $f_n$ is the normal force. In order to calculate $\tau_{max}$, we assume a uniform pressure distribution and define $r$ with respect to the center of mass ($r=r_{CM}$): $\tau_{max} = -\mu_s\frac{mg}{A} \int_{A}|r_{CM}|dA$. For quasi-static pushing, the velocity is aligned with the frictional load, and therefore must be parallel to the normal of the limit surface. This results in the following constraints on the object motion
\begin{equation}
\frac{v_x} {\omega} = c^2 \frac{f_x}{\tau}, \quad
\frac{v_y} {\omega} = c^2 \frac{f_y}{\tau}, \quad \text{and} \quad
c = \frac{\tau_{max}}{f_{max}}
\label{eq_quasi_static}
\end{equation}
used within our estimation factor graph in the next section.

\section{State Estimation with Factor Graphs}

\begin{figure*}[!t]
	\centering
	\captionsetup[subfigure]{justification=centering}
	\begin{subfigure}[b]{0.285\linewidth}
		\centering
		\includegraphics[scale=0.58]{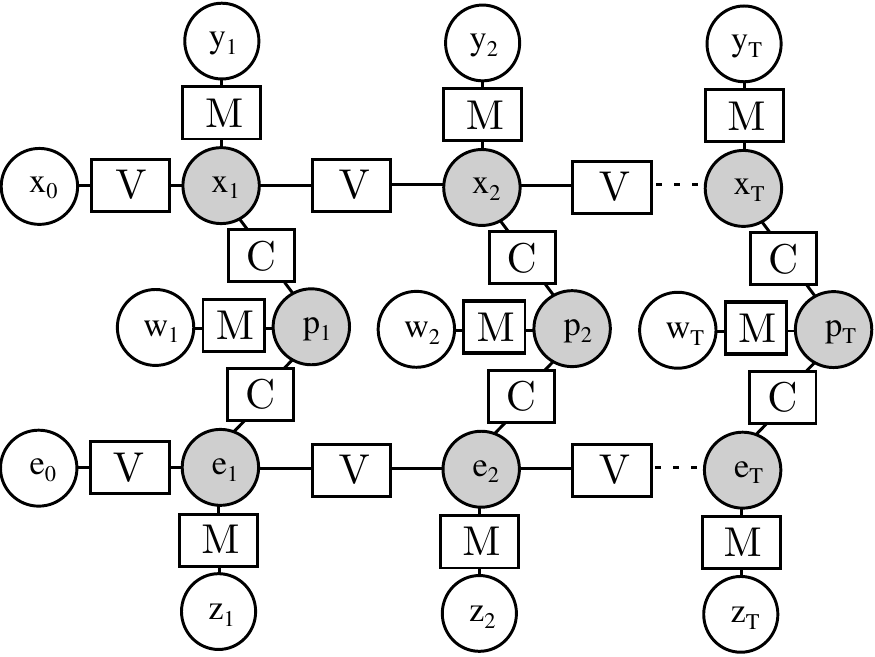}
		\caption{\small CP (pose + contact point meas.)}
		\label{CP_model}
	\end{subfigure}
	\hfill\vline\hfill
	\begin{subfigure}[b]{0.35\linewidth}
		\centering
		\includegraphics[scale=0.58]{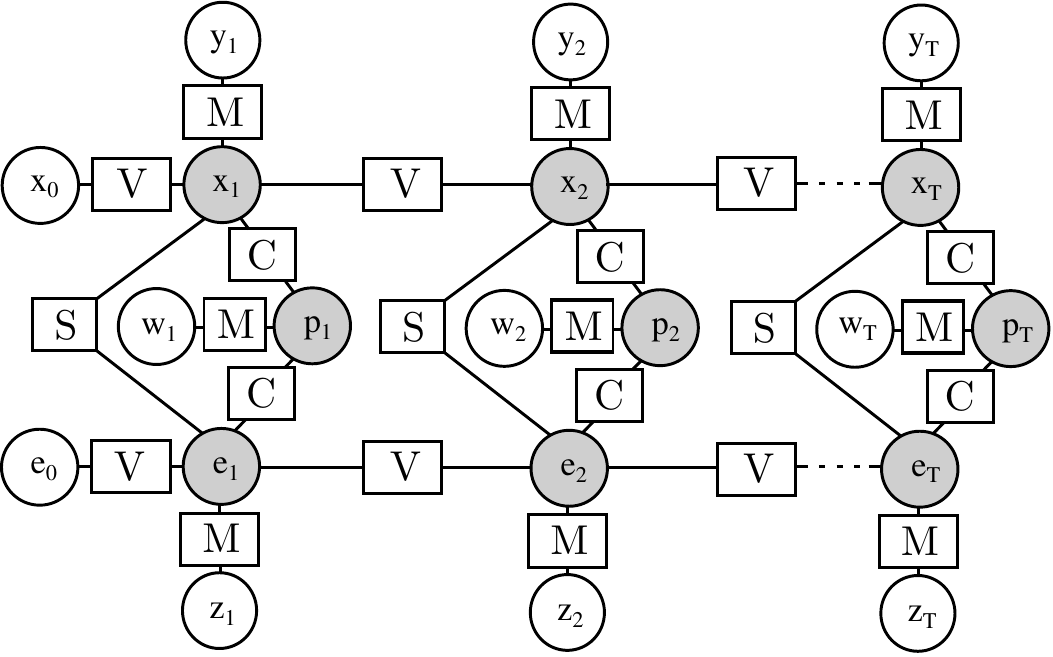}
		\caption{\small SDF (Intersection prior + CP)\\}
		\label{SDF_model}
	\end{subfigure}
	\hfill\vline\hfill
	\begin{subfigure}[b]{0.35\linewidth}
		\centering
		\includegraphics[scale=0.58]{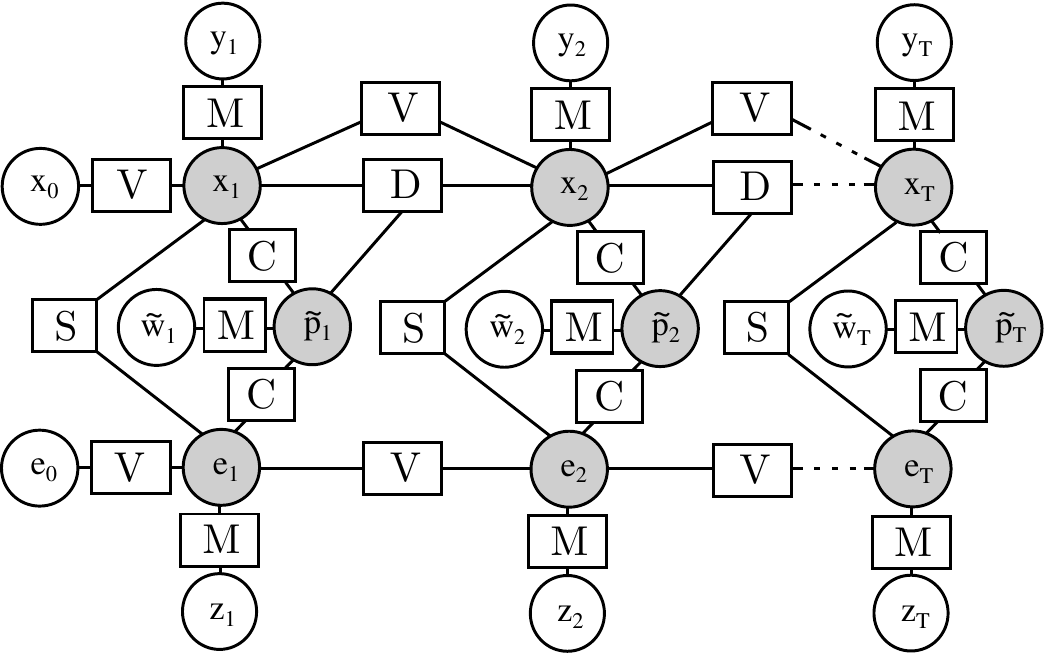}
		\caption{ \small QS (Quasi-static dynamics + SDF)}
		\label{QS_model}
	\end{subfigure}
	\caption{\small Estimation graphs. Filled circles are unknown state variables, unfilled circles are measured values, and squares indicate factors.}
	\label{fig:factor_graphs}
	\vspace{-4mm}
\end{figure*}

To solve state estimation during manipulation we formulate a factor graph of belief distributions over any state and force vector trajectory and perform inference over the trajectory given noisy sensor measurements. The graph construction and inference is performed with GTSAM~\cite{dellaert2006square,dellaert2012factor} via sparsity exploiting nonlinear least squares optimization to find the maximum a posteriori (MAP) trajectory that satisfies all the constraints and measurements. In the batch setting we use a Gauss-Newton optimizer and in an incremental setting we use iSAM2 that performs incremental inference via Bayes trees~\cite{kaess2012isam2}.
All random variables and measurements are assumed to have a Gaussian distribution. In the remainder of this section, we describe the construction of the relevant factor graphs depicted in Fig.~\ref{fig:factor_graphs}.

\subsection{Model Design}

We construct three different factor graphs for state estimation in our pushing task: CP, SDF, and QS (see Fig.~\ref{fig:factor_graphs}). All three models include the latent state variables for a given time $t$: the planar object pose $\vec{x}_t\in{SE(2)}$, the projected end-effector pose $\vec{e}_t\in{SE(2)}$, and the contact point $\vec{p}_t \in\mathbb{R}^2$.

\textbf{Measurements}: Each of the latent state variable is accompanied by an associated measurement factor $\mathrm{M}$ which projects corresponding measurements from $SE(3)$ into the pushing plane. The object poses are estimated by the visual tracking system with measurements $\vec{y}_t\in{SE(3)}$.  Likewise, the end-effector pose measurements $\vec{z}_t\in{SE(3)}$ may be provided from robot forward kinematics, or from the tracking system (DART includes a prior on joint measurements). The contact-point measurements $\vec{w}_t\in{SE(3)}$ are provided by a tactile sensor model. In the QS graph (Fig.~\ref{QS_model}), we include a new state variable for the applied planar contact force $\vec{f}_t\in\mathbb{R}^2$ with corresponding measurements $\vec{\alpha}_t\in\mathbb{R}^3$. For simplicity of graphical representation, we combine the contact point and force variables:
\begin{align}
\tilde{\vec{p}}_t = \begin{bmatrix} 
\vec{p}_t \\
\vec{f}_t
\end{bmatrix}, \quad
\tilde{\vec{w}}_t = \begin{bmatrix} 
\vec{w}_t \\
\vec{\alpha}_t
\end{bmatrix}
\end{align}

\textbf{Geometric Constraints}: We assume constant point-contact between the end-effector and the object. We include the factor $\mathrm{C}$ which incurs a cost on the difference between the contact point $\vec{p}_t$ and the closest point to a surface ($\xi$)~:
\begin{align}
\mathrm{C}(\xi,~\vec{p}_t) &= G\left(\xi, ~\vec{p}_t\right) - \vec{p}_t
\end{align}
where $G\left(\xi, ~\vec{p}_t\right)$ is the projection of $\vec{p}_t$ onto $\xi$, and $\xi=\xi\left(\cdot\right)$ returns the surface geometry of a body with a given pose: $\xi=\xi\left(\vec{x}_t\right)$ for the object, and $\xi=\xi\left(\vec{e}_t\right)$ for the end-effector.
Additionally, the object and the end-effector must be prevented from occupying the same region in space. Such a constraint is necessary in practice where contact-point estimation is often noisy. Therefore, we introduce a factor $\mathrm{S}$ to penalize intersecting geometries with a signed distance field. Let the point on the end-effector furthest into the object be denoted by $\vec{\delta}\in\mathbb{R}^2$, where $\vec{\delta}=\vec{\delta}\left(\vec{x},\xi\left(\vec{e}\right)\right)$. The projection of $\delta$ onto $\xi\left(\vec{x}\right)$ (the surface of the object) is then defined by $G_{\delta}=G\left(\vec{\xi\left(x\right)},\delta\right)$, and we can apply a penalty
\begin{align*}
\mathrm{S}(\vec{x},\vec{e}) = 
\begin{cases}
G_{\delta} - \vec{\delta}, &\quad\text{if intersecting}  \\
0, &\quad \mathrm{otherwise}
\end{cases}
\end{align*}

\textbf{Dynamics}: We add a constant velocity prior $V$ to impose smoothness on state transitions. For example, for finite-difference velocities of object poses we have
\begin{align}
V(\vec{x}_{t-1},\vec{x}_t,\vec{x}_{t+1}) = \frac{\vec{x}_t - \vec{x}_{t-1}}{\Delta_t} - \frac{\vec{x}_{t+1} - \vec{x}_t}{\Delta_{t+1}} 
\end{align}
where $\Delta_t$ and $\Delta_{t+1}$ denote the timestep sizes at $t$ and $t+1$.
Similar to \cite{yu2017realtime}, we introduce an additional factor $D$ to condition object state transitions on quasi-static pushing. The corresponding graphical model is denoted by QS and is shown in Fig. \ref{QS_model}. From Eq.~\ref{eq_quasi_static} we get
\begin{align}
D(\vec{x}_{t-1},\vec{x}_t,\tilde{\vec{p}}_t) = \frac{\vec{v}_t} {\omega_t} - c^2 \frac{\vec{f}_t}{\tau_t}
\end{align}
where $\vec{v}_t = ({\vec{x}_{\mathrm{trans},t} - \vec{x}_{\mathrm{trans},t-1})/\Delta_t}$ and $\omega_t=({\vec{x}_{\mathrm{rot},t} - \vec{x}_{\mathrm{rot},t-1})/\Delta_{t-1}}$ are the finite-difference linear and angular velocity, respectively.
The final cost function is optimized with respect to the set of variables $\Phi=\{  (\vec{x},\vec{e},\tilde{\vec{p}}) \}_{t=1}^{t=T}$ over a trajectory of length $T$ :
\begin{equation*}
\small
\begin{split}
\Phi^*&=\underset{\Phi}{\operatorname{arg\,min}}\sum^T_{t=1}
\bigg\{
\norm{\mathrm{D}(\vec{x}_{t-1},\vec{x}_t,\tilde{\vec{p}}_t))}_{\Sigma_{\mathrm{D}}}^2 
+ \norm{\mathrm{V}(\vec{x}_{t-1},\vec{x}_t,\vec{x}_{t+1})}_{\Sigma_{\mathrm{V}}}^2\\
&+ \norm{\mathrm{V}(\vec{e}_{t-1},\vec{e}_t,\vec{e}_{t+1})}_{\Sigma_{\mathrm{V}}}^2
+ \norm{\mathrm{C}(\vec{x}_t,\vec{e}_t)}_{\Sigma_{C}}^2
+ \norm{C(\tilde{\vec{p}}_t,\vec{x}_t)}_{\Sigma_{C}}^2 \\
&+ \norm{C(\tilde{\vec{p}}_t,\vec{e}_t)}_{\Sigma_{C}}^2
+ \norm{S(\vec{x}_t,\vec{e}_t)}_{\Sigma_{S}}^2
+ \norm{M(\vec{x}_t,\vec{y}_t)}_{\Sigma_{M}}^2 \\
&+ \norm{M(\vec{e}_t,\vec{z}_t)}_{\Sigma_{M}}^2 
+ \norm{M(\tilde{\vec{p}}_t,\tilde{\vec{w}}_t)}_{\Sigma_{M}}^2
\bigg\}
\end{split}
\end{equation*}
The above equation provides the locally optimal i.e. MAP solution of the estimation problem.

\section{Baseline Comparison }

In order to first ascertain the general performance of our approach, we evaluate the QS-graph on the MIT planar pushing dataset \cite{mitdata} using batch optimization. This data contains a variety of pushing trajectories for a single-point robotic pushing system. The object poses were tracked with a motion capture system, and contact forces were measured with a pushing probe mounted on a force-torque sensor. We use this data as ground truth, since it is considered to be sufficiently reliable. We restrict our experiments to a subset of this data, using trajectories with zero pushing acceleration and velocities under 10 cm/s in order to maintain approximately quasi-static conditions. Additionally, we only consider trajectories on the ABS surface, but examine different object types (ellip1, rect1, rect3) with approximately 100 trajectories per object and measurements provided at 100Hz. Gaussian noise is artificially added to the measurements prior to inference, with the following sigma values: $\sigma_{\vec{x}_{\mathrm{trans}}}=0.5\mathrm{cm},\; \sigma_{\vec{x}_{\mathrm{rot}}}=0.5 \mathrm{rad},\; \sigma_{\vec{e}_{\mathrm{trans}}}=0.5\mathrm{cm},\; \sigma_{\vec{e}_{\mathrm{rot}}}=0.5 \mathrm{rad},\; \sigma_{\vec{p}}=0.5\mathrm{cm},\; \sigma_{\vec{f}}=0.5\mathrm{N}$. 

\begin{figure}[t]
	\centering
	\includegraphics[width=0.99\linewidth]{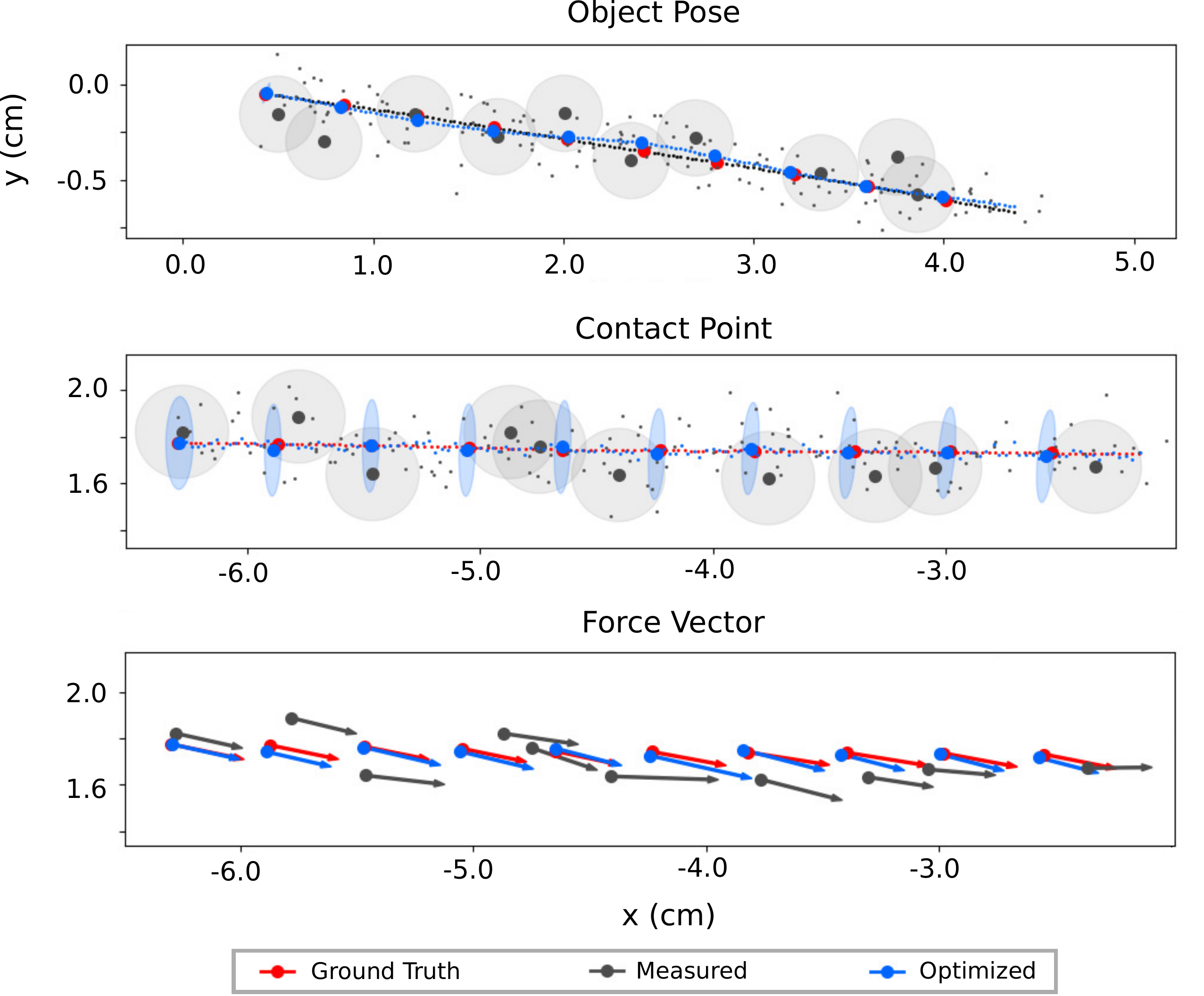}
	\caption{\small Example of performing the inference on a trajectory from the MIT pushing dataset, using the QS graph. Noise is artificially added to measurements prior to smoothing. Two-sigma contours and force vectors are displayed at every 15th time-step for visual clarity.}
	\label{traj1}
	\vspace{-4mm}
\end{figure}

The resulting RMS and covariance values post-optimization are shown in Table~\ref{big_table}. The optimized values exhibit marked reductions in error compared to the sigma values of the initial measurements. Note that, for object poses we only include values in which the object is in motion, in order to exclude trivial stationary estimates. All position-related values are in cm, with angular values in radians, and forces in Newtons. An example of an optimized trajectory is shown in Fig.~\ref{traj1}. Although the observation noise is artificial, these results indicate that latent state estimates may still be successfully recovered with the addition of geometric and physics-based priors, and without over-constraining the optimization. 

\begin{table}[!t]
\begin{center}
	\caption{\small RMS and Covariance values on the MIT Dataset.}
	\label{big_table}
	\begin{subtable}{\linewidth}
	\centering
	\scalebox{1.}{
	\begin{tabular}{c|*4c}
		\hline
		Object  &  $\text{RMS}\left(\vec{x}_{\mathrm{trans}}\right)$  &
		 		   $\text{RMS}\left(\vec{x}_{\mathrm{rot}}\right)$ &
		 		   $\Sigma\left({\vec{x}_{\mathrm{trans}}}\right)$ &
		 		   $\Sigma\left({\vec{x}_{\mathrm{rot}}}\right)$ \\
		\hline
		ellip1  &  0.0262 & 0.283    & 2.723e-4 & 4.171e-10 \\
		rect1   &  0.0253 & 3.471-5  & 2.931e-4 & 4.19e-10 \\
		rect3   &  0.0182 & 1.672e-5 & 2.563e-4 & 4.18e-10 \\
		\hline 
		Object  &  $\text{RMS}\left(\vec{e}_{\mathrm{trans}}\right)$  &
		$\text{RMS}\left(\vec{e}_{\mathrm{rot}}\right)$ &
		$\Sigma\left({\vec{e}_{\mathrm{trans}}}\right)$ &
		$\Sigma\left({\vec{e}_{\mathrm{rot}}}\right)$   \\
		\hline
		ellip1  & 7.73e-2 & 9.47e-2 & 4.74e-3 & 7.11e-3 \\
		rect1   & 8.59e-2 & 9.18e-2 & 5.89e-3 & 6.01e-3 \\ 
		rect3   &  0.372 & 0.376 & 0.148 & 0.154 \\
		\hline
		Object  &  $\text{RMS}\left(\norm{\vec{f}}\right)$  & 
		$\text{RMS}\left(\vec{f}_{\mathrm{rot}}\right)$  &
		$\Sigma\left(\norm{\vec{f}}\right)$ &
		$\Sigma\left(\vec{f}_{\mathrm{rot}}\right)$ \\
		\hline
		ellip1  &  0.118 & 9.543e-2 & 9.827e-3 & 1.635e-4 \\ 
		rect1   &  0.145 & 9.683e-2 & 9.862e-3 & 1.823e-4 \\
		rect3   &  0.113 & 9.754e-2 & 9.145e-3 & 1.856e-4 \\
		\hline 
		Object  &  $\text{RMS}\left(\vec{p}_{\mathrm{trans}}\right)$  & 
		   --- & 
		$\Sigma\left(\vec{p}_{\mathrm{trans}}\right)$ &
		  --- \\
		\hline
		ellip1  &  3.42e-2 & --- & 2.54e-3 & --- \\
		rect1   &  4.52e-2 & --- & 6.21e-3 & --- \\
		rect3   &  3.26e-2 & --- & 3.41e-3 & --- \\
		\hline
	\end{tabular}}
	\end{subtable}
\end{center}
\end{table}

\section{State Estimation in Open and Cluttered scenes}

\begin{figure}[!t]
	\centering
	\includegraphics[width=1.02\linewidth]{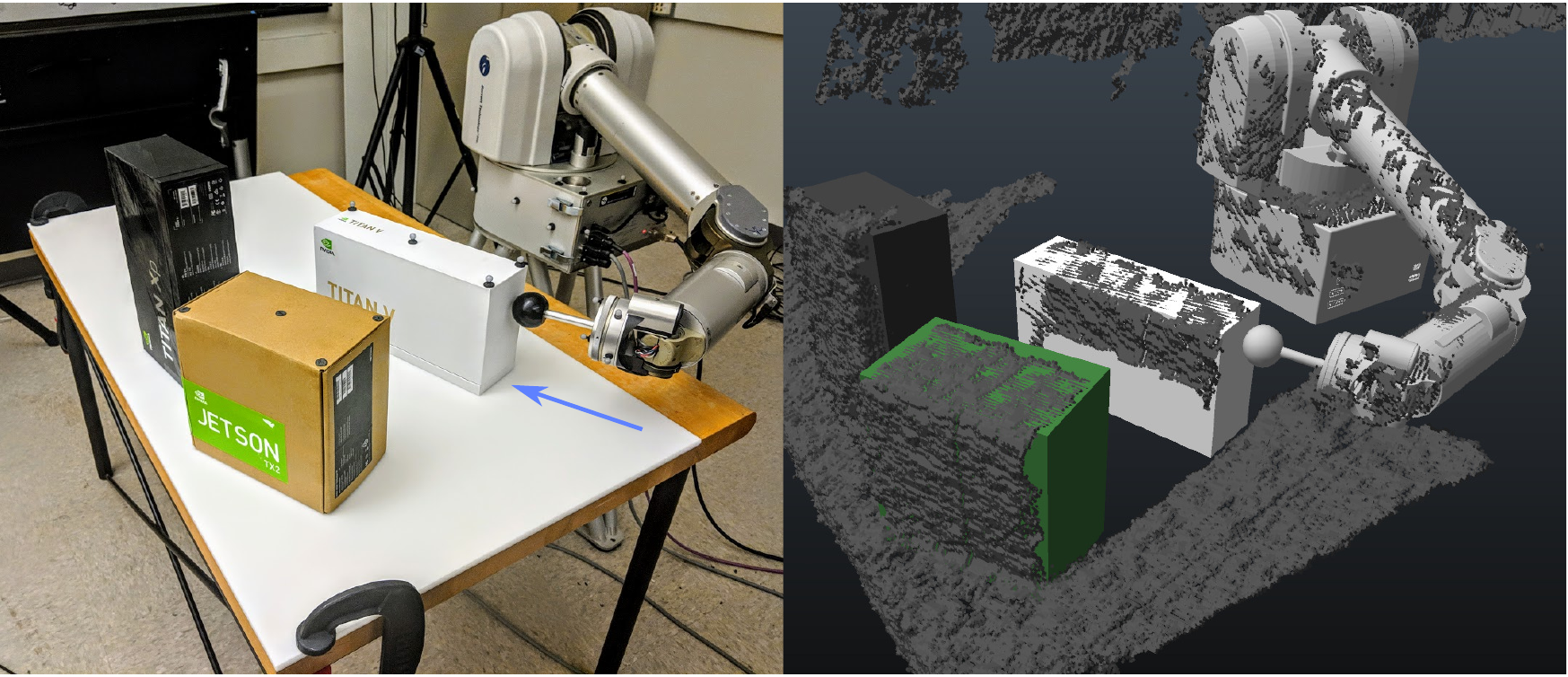}
	\caption{\small Left: Setup for pushing experiments with occlusion using Barrett-WAM manipulator. The white box is the pushed object, with general pushing direction indicated by the blue arrow. The system is observed by a depth camera to the left (out of frame). Right: visualization of the tracked system in DART \cite{schmidt2014dart}, with the observed pointcloud marked in dark grey.}
	\label{robot_setups}
\end{figure}

We first perform pushing experiments with the Barrett WAM manipulator acting on a laminated box as shown in Fig.~\ref{robot_setups}. The system is observed by a stationary PrimeSense depth camera located 2.0m away from the starting push position of the end-effector. Vision-based tracking measurements of the object pose are provided by DART, configured with contact-based priors and joint estimates~\cite{schmidt2015depth}. The robot is equipped with a Force-Torque sensor and a rigid end-effector mounted with a spherical hard-plastic pushing probe. The contact forces are measured by the F/T sensor, with contact point measurements provided through optimization in DART. Ground-truth poses are provided via a motion-capture system. The table is mounted with a smooth delrin sheet to provide approximately uniform friction across the pushing area.  

\begin{figure}[!t]
	\centering
	\begin{subfigure}{0.9\linewidth}
		\includegraphics[width=\linewidth]{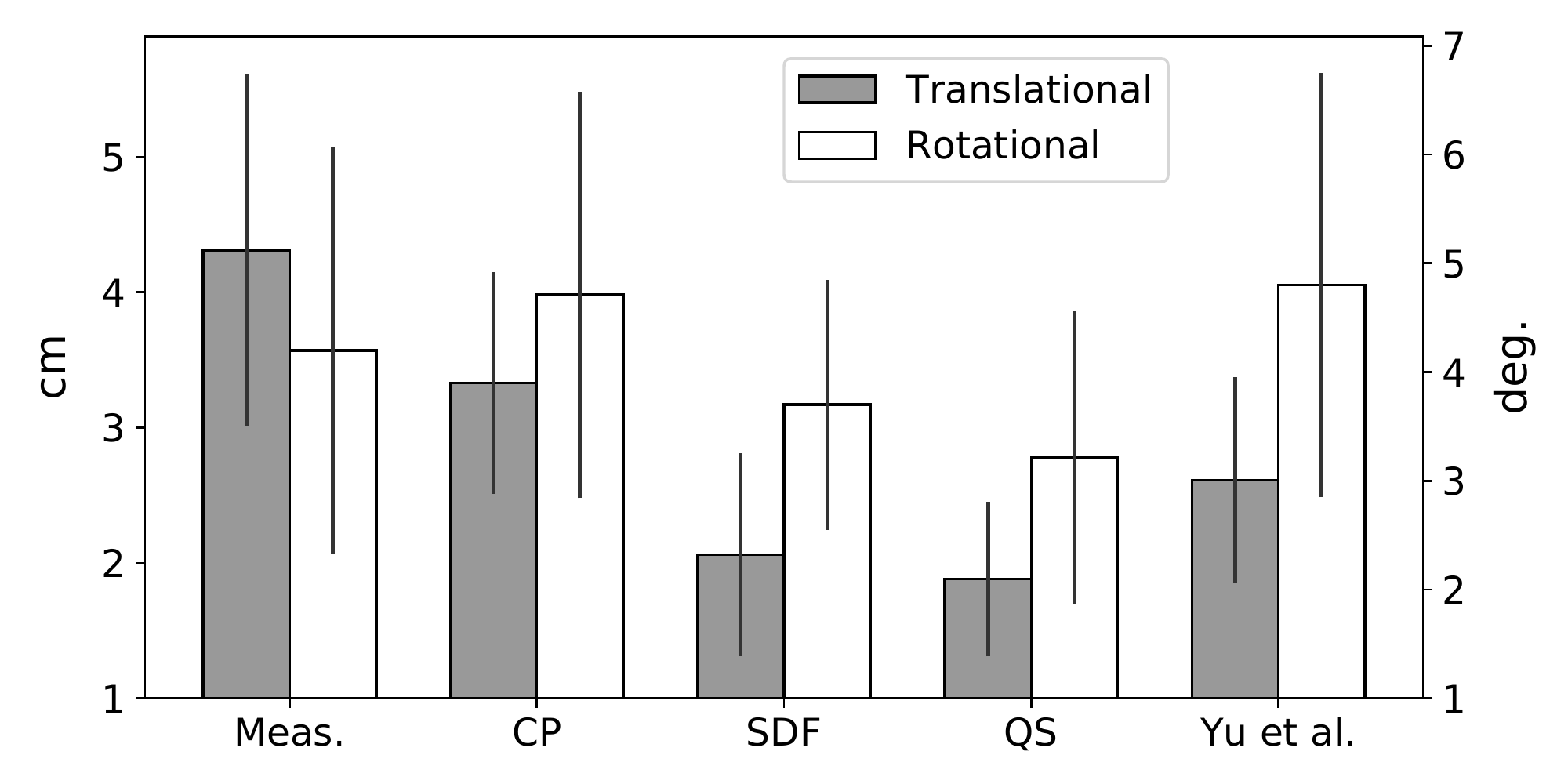}
		\caption{Fully observable}
		\label{fig:wam_clear}
	\end{subfigure}
	\begin{subfigure}{0.9\linewidth}
		\includegraphics[width=\linewidth]{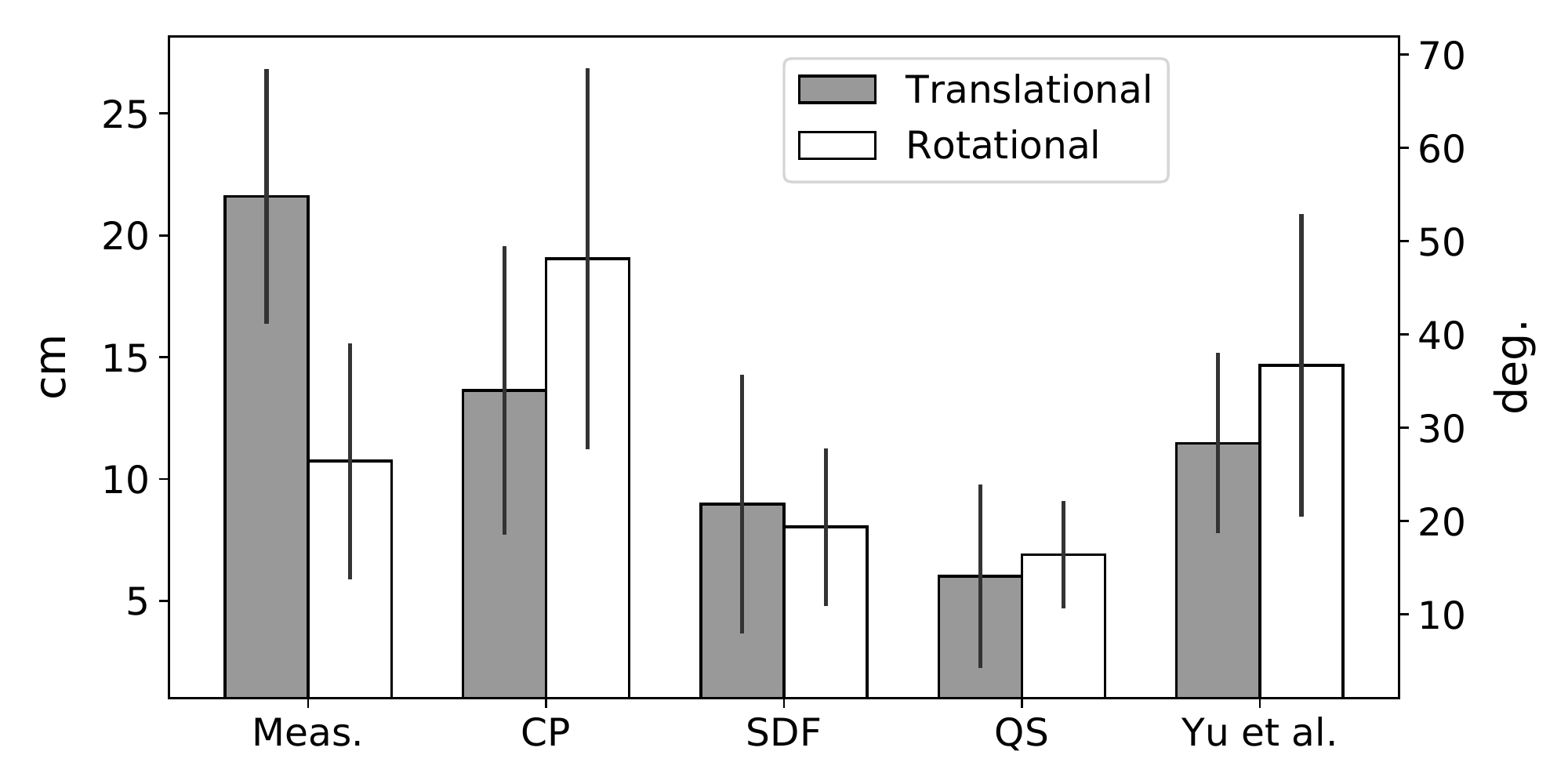}
		\caption{Occluded}
		\label{fig:wam_clear}
	\end{subfigure}
	\caption{\small Mean error and standard deviations of object pose estimates (after the last iSAM2 step has been performed). CP, SDF, and QS model results are compared raw measured values, and to those produced by the graph described in Yu et al. \cite{yu2017realtime}. Tracking performance is greatly improved with the inclusion of geometric and physics-based priors. The comparison with \cite{yu2017realtime}, which does not use SDF priors, indicates the importance of enforcing these constraints in practice.}
	\label{fig:wam_error_barchart}
	\vspace{-4mm}
\end{figure}

\begin{figure*}[!t]
	\centering
	\begin{subfigure}[!t]{0.475\linewidth}
		\centering
		\includegraphics[width=\linewidth]{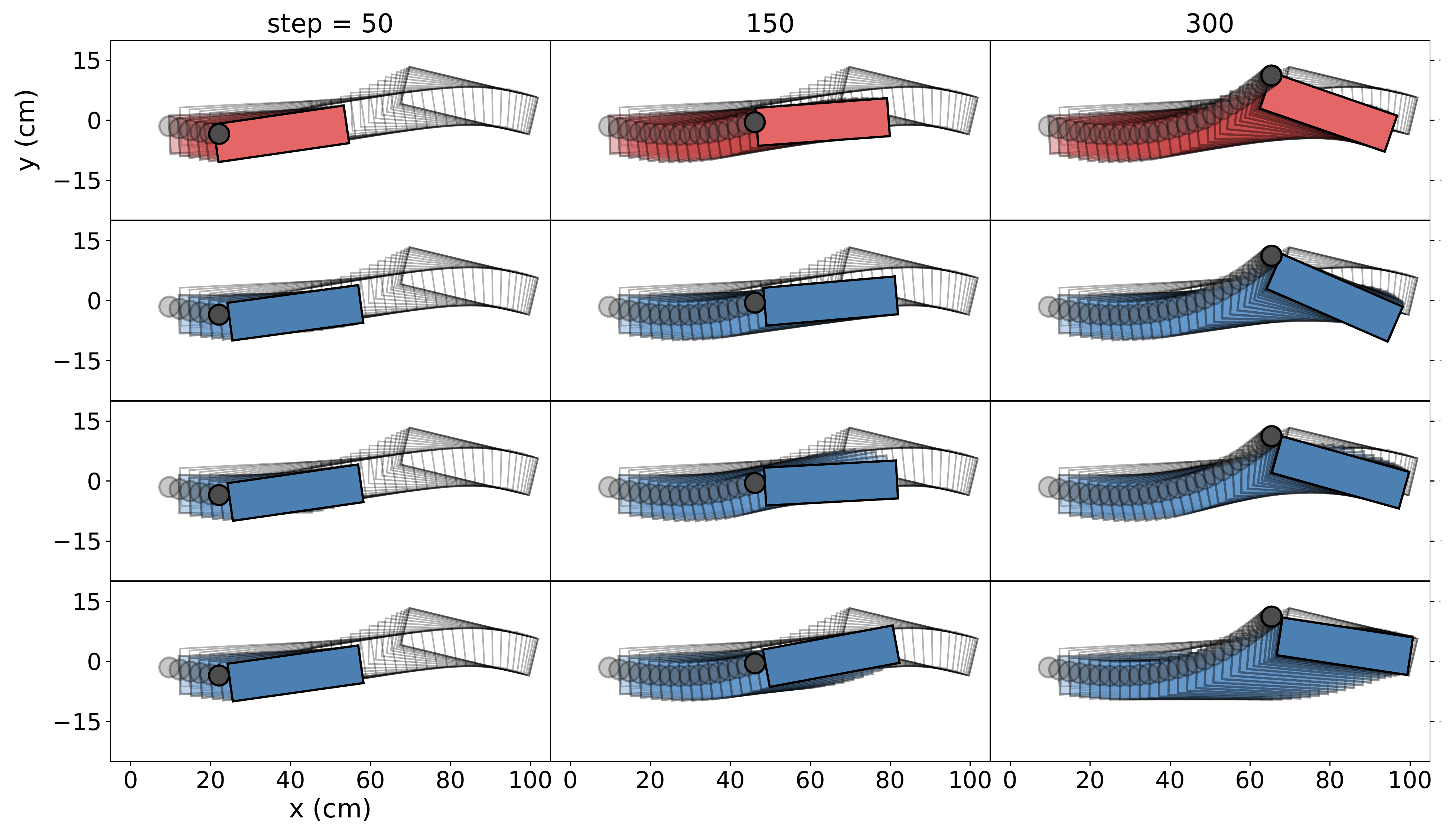}
		\caption{Trajectory 1 (fully observable)}
		\label{fig:wam_clear}
	\end{subfigure}
\hspace{-3mm}
	\begin{subfigure}[!t]{0.52\linewidth}
		\centering
		\includegraphics[width=\linewidth]{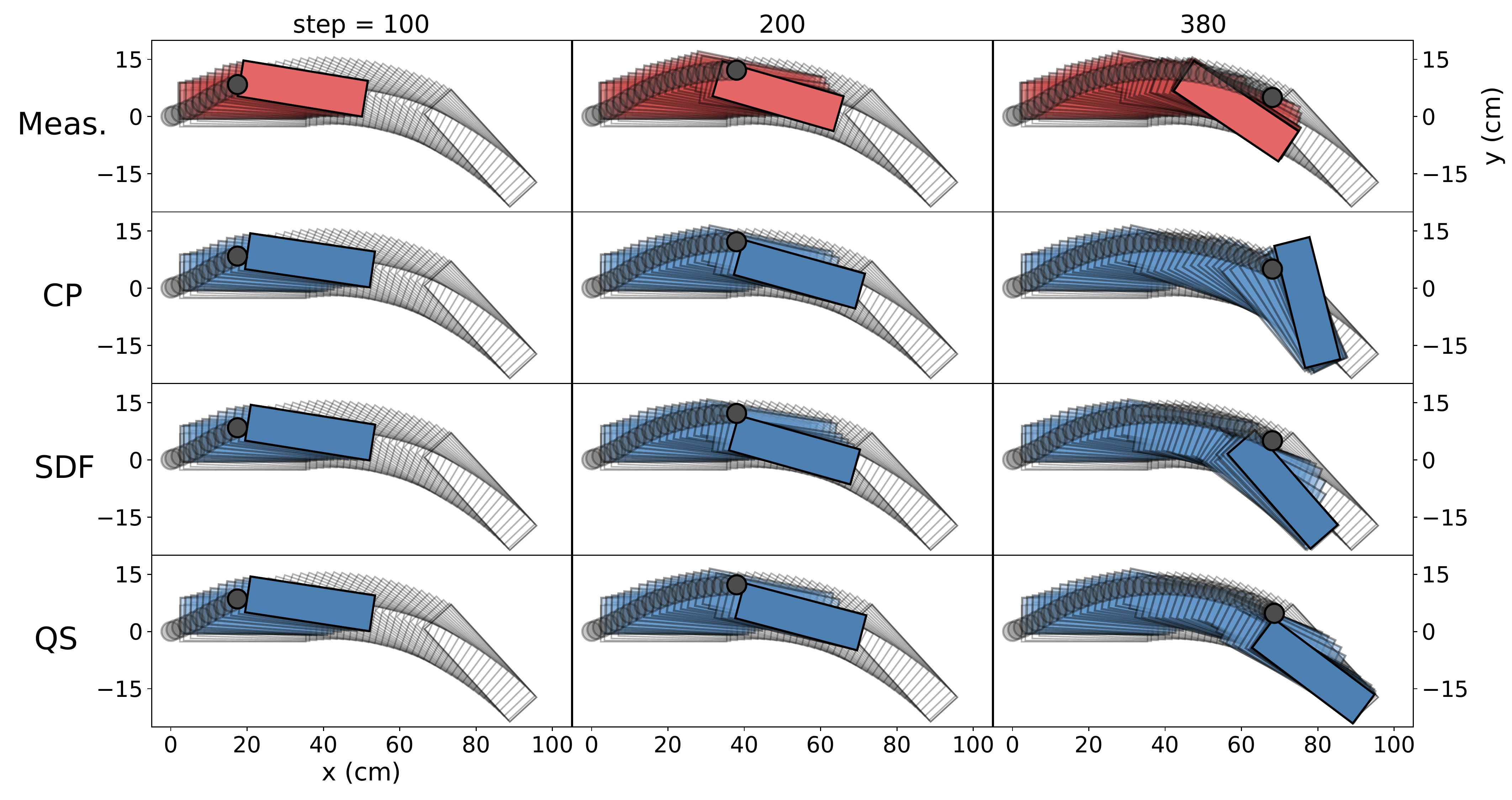}
		\caption{Trajectory 2 (with occlusion)}
		\label{fig:wam_occluded}
	\end{subfigure}
	\caption{\small Examples of estimated object trajectories for both un-occluded and occluded scenarios. Measured object pose histories (pink) are shown in the top rows, and compared below to the incrementally-optimized trajectories (blue) using the CP, SDF, and QS factor graphs illustrated in Fig.~\ref{fig:factor_graphs}. Each column depicts the state estimates at a particular timestep (with respect to object pose measurements). The trajectories are overlayed onto the full ground-truth trajectories derived from motion-capture, with every 10 timestep intervals shown. Trajectories of the end-effector (grey circle) are also represented. The measurements show how the tracking system performance degrades under certain orientations, since less of the object is ``seen" as it turns away from the camera. Occlusion causes the system to lose track of the object entirely. Contact-point factors are insufficient for reliable tracking, and can cause object orientation to deviate wildly under occlusion. Incorporating SDF constraints helps to prevent many infeasible poses. The QS graph enforces pose changes which adhere to pushing mechanics. The physics-based priors inform the pose estimates, and stabilize the trajectory even under occlusion.}
	\label{fig:wam_results}
	\vspace{-2mm}
\end{figure*}

We performed 100 pushing trials with varying initial end-effector and object poses. The end-effector trajectories were varied in curvature and maintained a translational speed close to 6 cm/s to approximate quasi-static conditions. Object pose-tracking measurements were provided at roughly 25Hz, with end-effector poses and force/contact measurements published at 250Hz. Incremental inference of the factor graph is performed after 5 object pose measurements. 

In order to provide real-time performance, DART maintains a belief distribution over state at a single timestep. However, this can make tracking susceptible to unreliable measurements that may arise from state-dependent uncertainty or partial observability. As such, we purposely include trajectories in which the object orientation changes significantly with respect to the camera orientation, causing large variations in pointcloud association. In addition, the pushing trajectories were also performed in cluttered scenes, as depicted in Fig.~\ref{robot_setups}, with 85\% occlusion of the pushing object occurring in the middle of the trajectory.

Examples of measured and estimated state trajectories are shown in Fig.~\ref{fig:wam_results}. In the fully-observable (unocccluded) setting, distinct improvement of the object pose can be seen with both SDF and QS models. Under heavy occlusion, the visual tracking system loses the object and is unable to regain the trajectory state. However, the addition of both geometric and physics based priors to the factor graph result in realignment of the tracked object. Fig.~\ref{fig:wam_error_barchart} shows the tracking performance for fully observable trajectories using the CP, SDF, and QS factor graphs. The results are compared to the model proposed by Yu et al.~\cite{yu2017realtime}, which includes quasi-static dynamics factors with contact and zero-velocity priors.

\begin{table}[!t]
	\begin{center}
		\caption{\small Error Results for Force and Contact Recovery}
		\label{force_table}
		\begin{subtable}{\linewidth}
			\centering
			\scalebox{1.1}{
				\begin{tabular}{c|*3c}
					\hline
					Component  &  $\text{RMSE}$ &  $\text{MAE}$ &  $\sigma$\\
					\hline
					Force magnitude (N)   &  0.352  &  0.195 &  0.043 \\
					Force direction (deg.)&  3.15   & 2.54 &  0.78 \\
					Contact location (cm) & 0.32 &  0.14 &  0.18 \\
					\hline
			\end{tabular}}
		\end{subtable}		
	\end{center}
	\vspace{-5mm}
\end{table}

\begin{figure}[!t]
	\centering
	\includegraphics[width=0.90\linewidth]{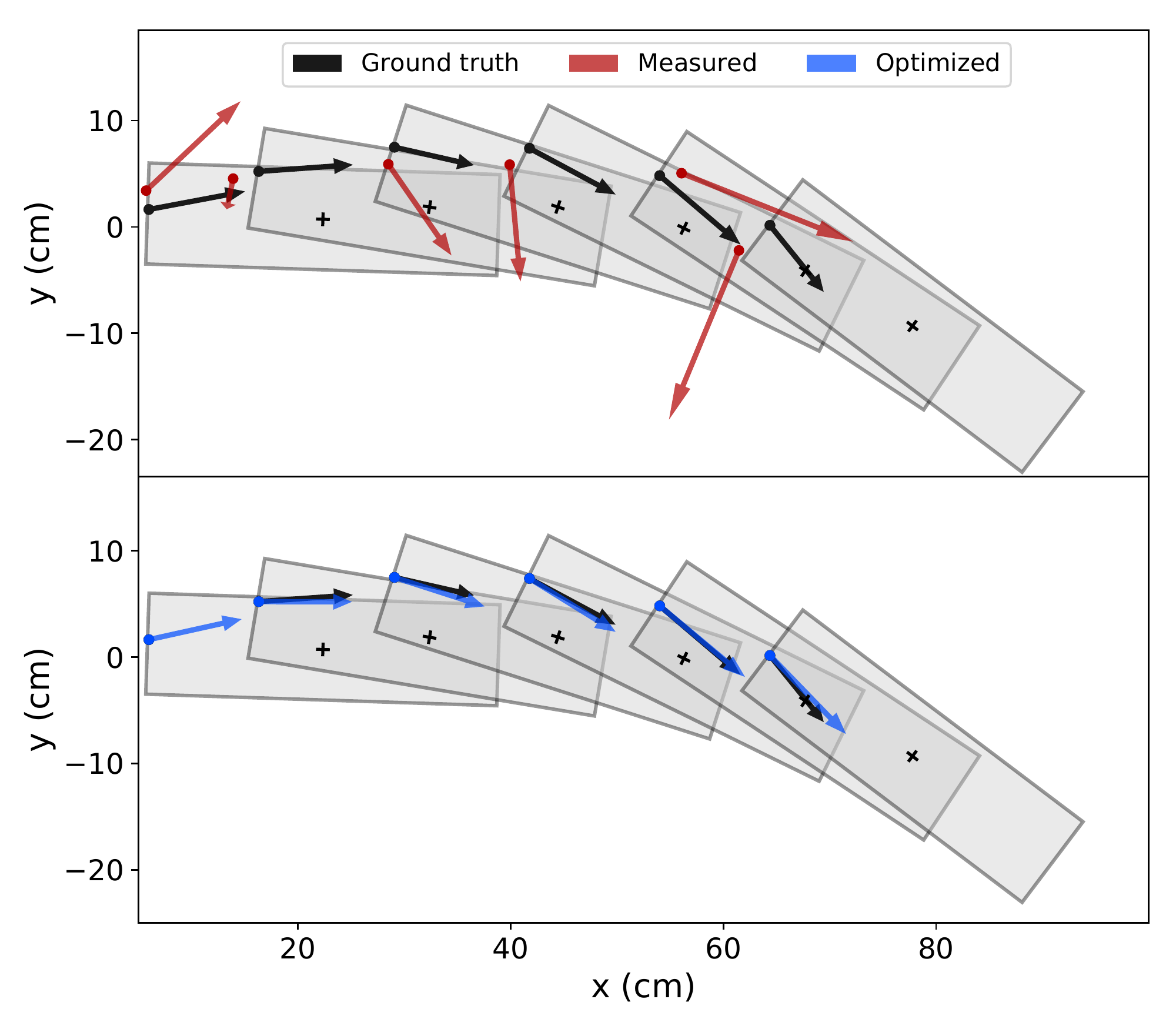}
	\caption{\small Example of force-estimation using the QS model with ground-truth poses and non-Gaussian noise added to force measurements and contact points. Force vectors and contact points are recovered by the optimization process.}
	\label{fig:force}
	\vspace{-4mm}
\end{figure}

In addition to improving inference on kinematic trajectories, the QS graph can be used to improve contact point and force estimates. To demonstrate this, we artificially add non-Gaussian noise (bi-modal mixture of two triangular distributions) to contact points and force measurements on the ground-truth data. The resulting estimation errors after optimization are shown in Table~\ref{force_table}, and indicate that our approach manages to recover true contact points and pushing forces. An an example of force-trajectory optimization is illustrated in Fig.~\ref{fig:force}.


\section{Force Estimation for Tactile Sensing}

We further demonstrate inference on force trajectories using realistic (noisy) tactile data. The Biotac sensor comprises of a solid core encased in an elastomeric skin and is filled with weakly-conductive gel~\cite{biotac}. The core surface is populated by an array of 19 electrodes, each measuring impedance as the thickness of the fluid between the electrode and the skin changes. A transducer provides static pressure readings which consist of a single scalar value per time-step. This sensor is also equipped with a thermistor for measuring fluid temperature. Although the device does not directly provide a force distribution or contact point measurements, an analytical method for estimating these values is described in~\cite{biotac}.  

Using an ABB YUMI robot with a mounted Biotac sensor, we generated randomized linear trajectories of the end effector pushing a 0.65 kg box across a laminated surface (see Fig.~\ref{intro_img}) starting from a number of different poses. We used the DART tracking system \cite{schmidt2015depth} to obtain object and end-effector pose measurements, along with approximate contact points. The analytical force sensor model~\cite{biotac}, was used to provide initial force measurements.

\begin{figure}[!t]
	\centering
	\begin{subfigure}[b]{0.95\linewidth}
		\centering
		\hspace{-8mm}
		\includegraphics[width=\linewidth]{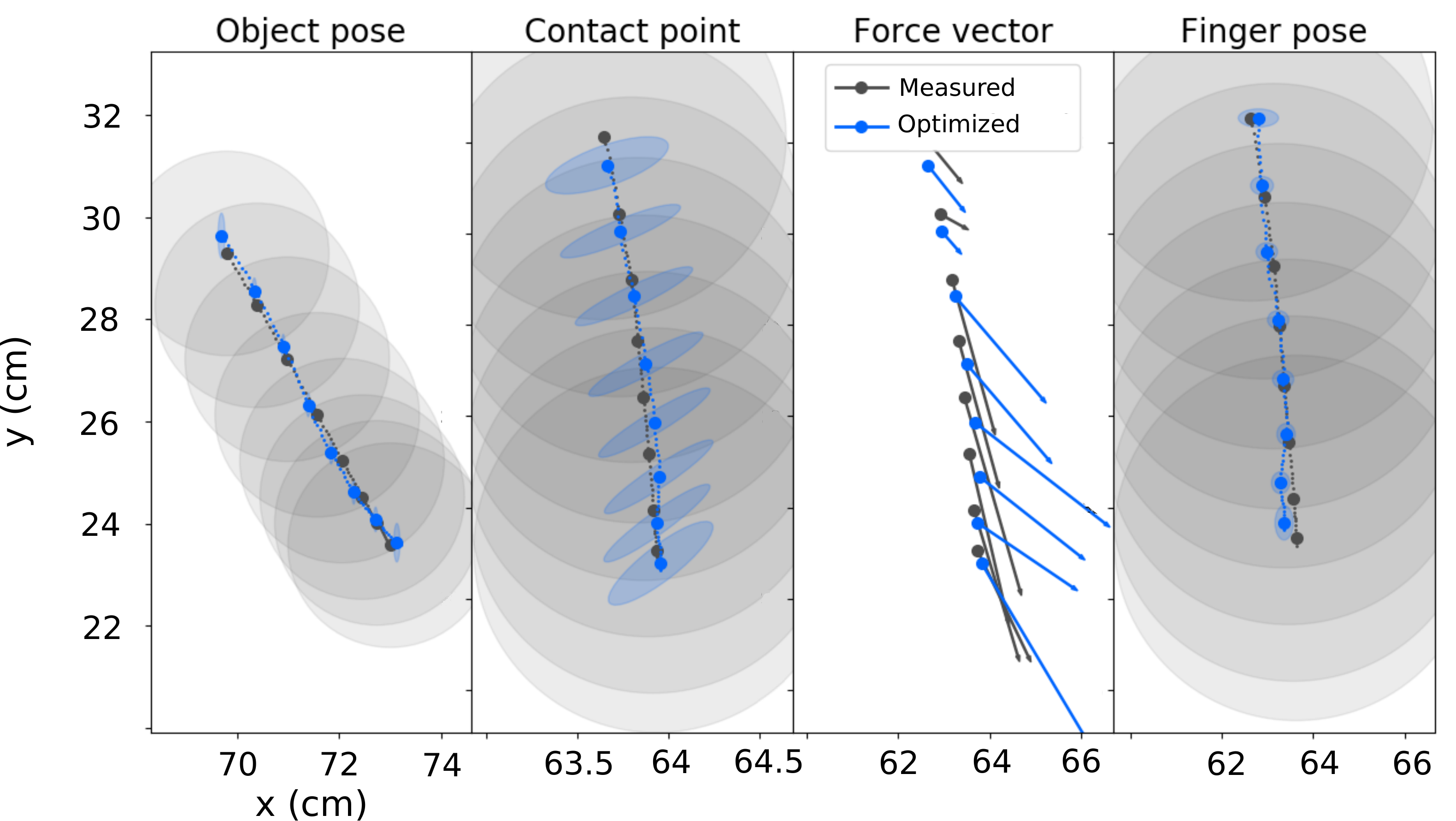}
		\caption{Trajectory 1}
		\label{bt_traj_1}
	\end{subfigure}
\quad
	\begin{subfigure}[b]{0.95\linewidth}
		\centering
		\hspace{-8mm}
		\includegraphics[width=\linewidth]{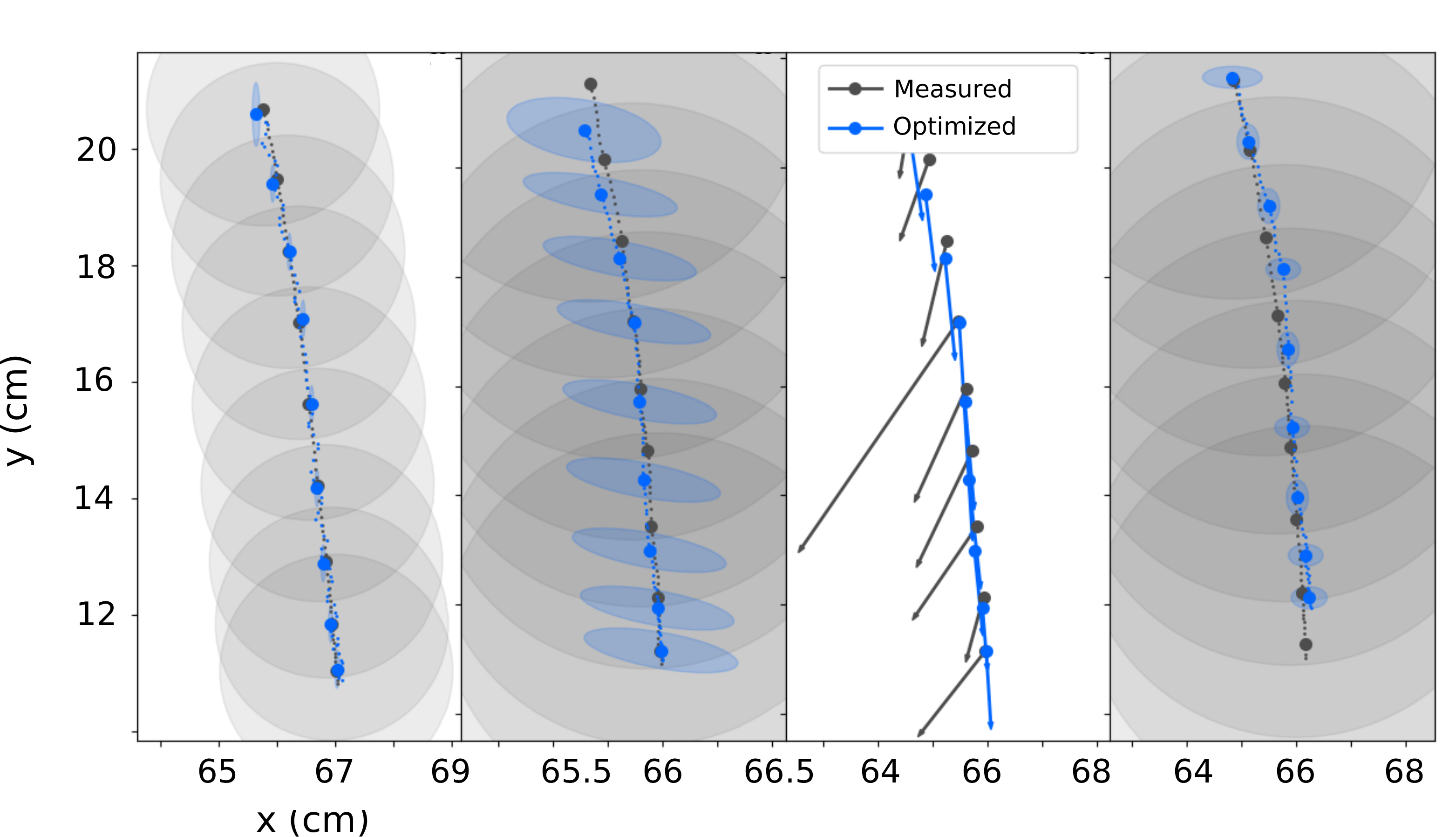}
		\caption{Trajectory 2}
		\label{bt_traj_2}
	\end{subfigure}
	\caption{\small Examples of pushing trajectories performed on the YUMI system. Initial object and finger pose estimates are provided by the DART tracking system. Contact points and force measurements are estimated by the analytic tactile sensor model \cite{biotac}. Each trajectory is optimized using the QS graph depicted in Fig.~\ref{QS_model}. Two-sigma values and force vectors shown at every 10th timestep for visual clarity. Joint inference over kinematic and force trajectories decreases uncertainty in poses as well as contact points and forces, and smoothens noisy tactile data to agree with physics-based constraints.} 
	\label{bt_traj_result}
	\vspace{-4mm}
\end{figure}

\begin{figure}[h]
	\centering
	\begin{subfigure}[b]{0.8\linewidth}
		\centering
		\includegraphics[width=\linewidth]{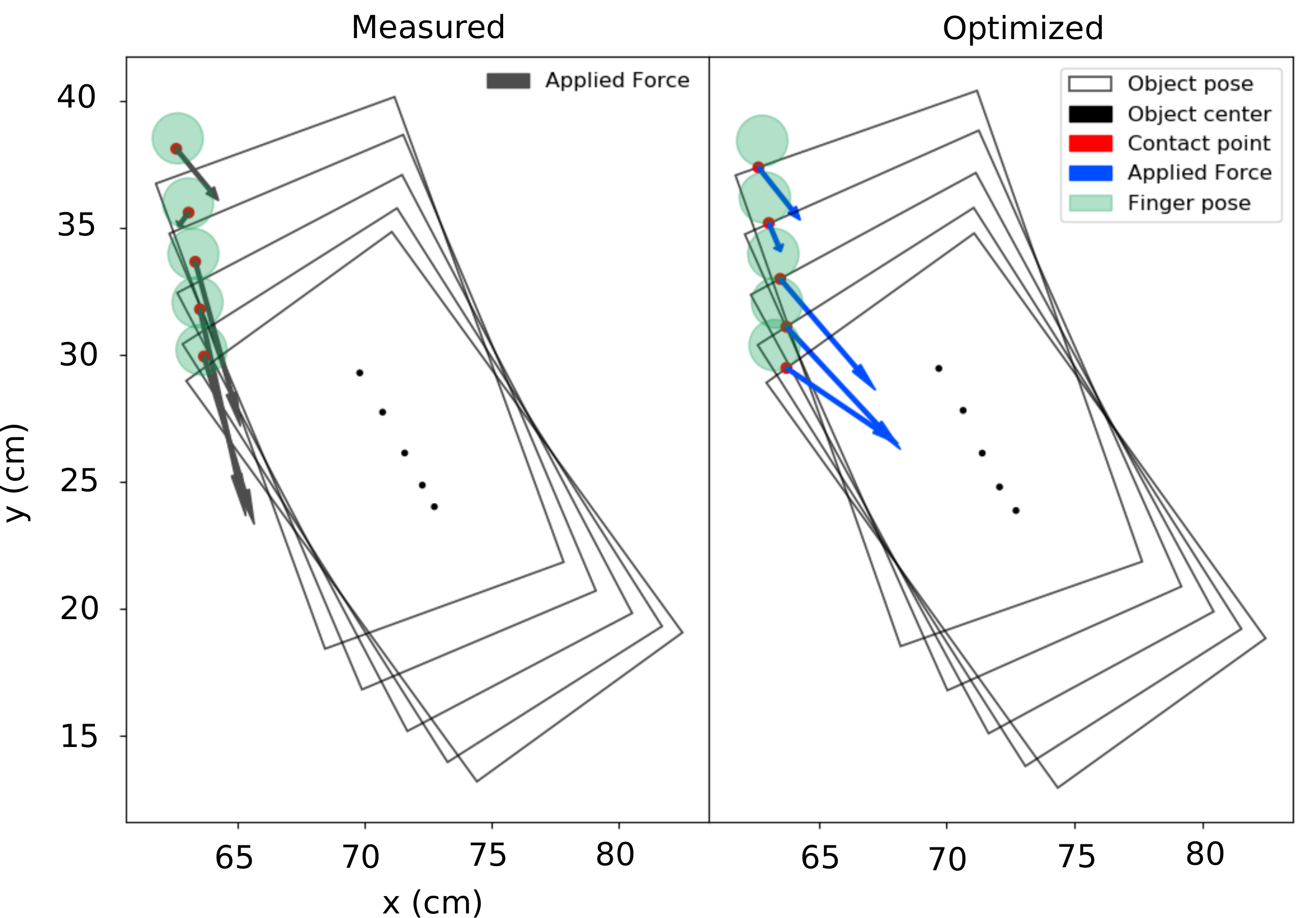}
		\caption{Trajectory 1}
		\label{bt_viz_1}
	\end{subfigure}
	\begin{subfigure}[b]{0.61\linewidth}
		\centering
		\includegraphics[width=\linewidth]{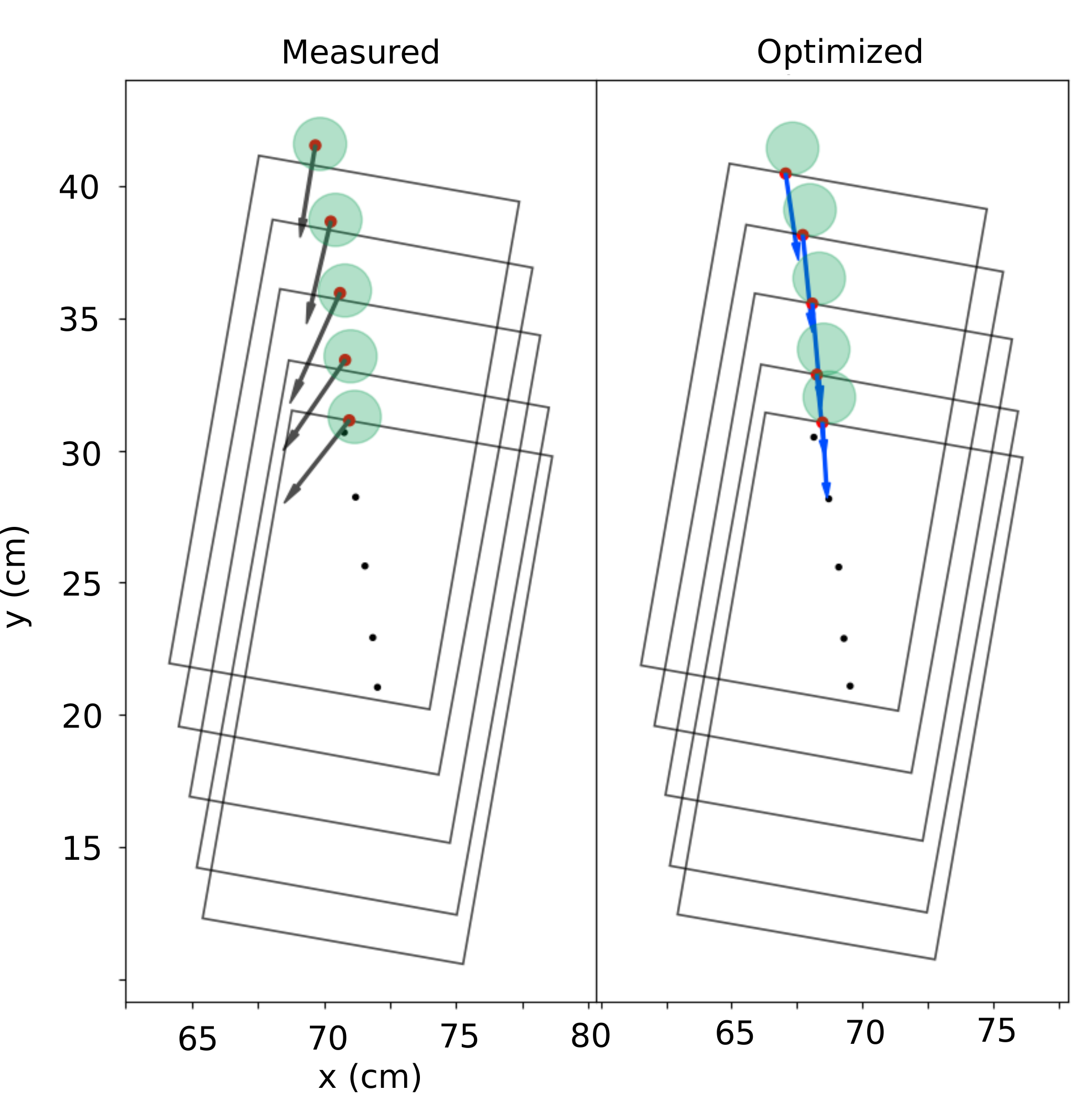}
		\caption{Trajectory 2}
		\label{bt_viz_2}
	\end{subfigure}
	\caption{\small Visualizations of measurements for corresponding trajectories in Fig.~\ref{bt_traj_result}. Measured positions, contact points and force-vector outputs from the learned sensor model are shown on the left-hand side. Optimized values are shown on the right, indicating consistency of finger-object surface contact. Our approach produces force trajectories which more closely adhere to quasi-static mechanics. Joint inference allows kinematic trajectories to inform the force estimates, aligning forces to the object center of mass during linear motion, and correcting applied moments when motion is non-linear.}
	\label{bt_raj_viz}
	\vspace{-4mm}
\end{figure}

Examples of initial and optimized trajectories are shown in Fig.~\ref{bt_traj_result}-\ref{bt_raj_viz}. The presence of the contact surface factor shrinks the contact point covariance in the direction of push, as is expected. The covariances for finger and object pose estimates are drastically reduced, exhibiting the benefits of joint-inference across trajectory histories. Also, the dynamics factor aligns the force vector in the direction of motion of the object. This is further clarified in Fig.~\ref{bt_raj_viz}, where force vectors are correctly aligned with the object center-of-mass for linear trajectories, and provide a moment arm during angular displacement. This demonstrates the importance of contact and geometric factors in aligning the surface tangents of the finger and the object at the point of contact.

\section{Conclusion}

We proposed a factor graph-based inference framework to solve estimation problems for robotic manipulation in batch and incremental settings. Our approach can leverage geometric and physics-based constraints along with vision and tactile based multi-modal sensor information to jointly estimate the history of robot and objects poses along with contact locations and force vectors. We perform several benchmarks on various datasets with multiple manipulators in real environments and show that our framework can contend with sensitive, noisy sensor data and occlusions in vision to efficiently solve for locally optimal state estimates that closely match ground truth. Future work will include incorporating the approach within a motion planning context \cite{Mukadam-RSS-17}, combining vision and tactile modalities in learning predictive sensor models \cite{lambert2018deep,sundaralingam2018robust}, and the possibility of integration into a hierarchical task-planning framework.


\newpage
\balance
\bibliographystyle{IEEEtran}

\end{document}